\documentclass[journal]{IEEEtran} %llncs}
\usepackage{amssymb}
\usepackage{graphicx}
\usepackage{textcomp}
\usepackage{amsmath}
\usepackage{color}
\usepackage{lineno}
\usepackage{url}
\usepackage{cite}
\usepackage{enumerate}
\usepackage{epstopdf}

\usepackage{graphicx}
\usepackage{graphics}
\usepackage{multirow}
\usepackage{caption}
\usepackage{subcaption}
\usepackage{gensymb}
\usepackage{float}
\usepackage[export]{adjustbox}

\usepackage{multirow}
\usepackage{array}
\usepackage{booktabs}
\usepackage{hhline}
\usepackage{xspace}
\usepackage{breqn}
\usepackage{tablefootnote}
\usepackage{verbatim}

\usepackage{algorithm}
\usepackage{algpseudocode}
\usepackage{soul}

\newcommand{\comst}[1]{}

\definecolor{orange}{rgb}{1,0.5,0}

\newcommand{\ours}{RSDNet}

\newcommand{\com}[1] {}

\newcommand{\miniTitle}[1]{\vspace{0.2cm}\noindent\textbf{#1}}

\graphicspath{{./figures/}}
\allowdisplaybreaks
\usepackage{array}
\usepackage{arydshln}
\usepackage{hyperref}

\begin{document}

\title{\ours: Learning to Predict Remaining Surgery Duration from Laparoscopic Videos Without Manual Annotations}

\author{Andru Putra Twinanda, Gaurav Yengera, Didier Mutter, Jacques Marescaux and Nicolas Padoy*

\thanks{This work was supported by French state funds managed within the Investissements d'Avenir program by the ANR (references ANR-11-LABX-0004 and ANR-10-IAHU-02) and by BPI France (project CONDOR). The authors would also like to acknowledge the support of NVIDIA with the donation of a GPU used in this research. \textit{Asterisk indicates corresponding author.}

Andru Putra Twinanda, Gaurav Yengera and Nicolas Padoy are with ICube, University of Strasbourg, CNRS, IHU Strasbourg, France (email: andru.putra@gmail.com, g.yengera@gmail.com, npadoy@unistra.fr)

Didier Mutter and Jacques Marescaux are with the University Hospital of Strasbourg, IRCAD, IHU Strasbourg, France.

Copyright (c) 2018 IEEE. Personal use of this material is permitted. However, permission to use this material for any other purposes must be obtained from the IEEE by sending a request to pubs-permissions@ieee.org.

The final version of this paper is A. P. Twinanda, G. Yengera, D. Mutter, J. Marescaux and N. Padoy, "RSDNet: Learning to Predict Remaining Surgery Duration from Laparoscopic Videos Without Manual Annotations," in IEEE Transactions on Medical Imaging. DOI: 10.1109/TMI.2018.2878055  Available at: \href{http://dx.doi.org/10.1109/TMI.2018.2878055}{http://dx.doi.org/10.1109/TMI.2018.2878055}
}}%

\markboth{}
{A. P. Twinanda \MakeLowercase{\textit{et al.}}: \ours: Prediction of Remaining Surgery Duration Without Manually Annotated Labels}
\maketitle

\begin{abstract}
Accurate surgery duration estimation is necessary for optimal OR planning, which plays an important role in patient comfort and safety as well as resource optimization. It is, however, challenging to preoperatively predict surgery duration since it varies significantly depending on the patient condition, surgeon skills, and intraoperative situation. In this paper, we propose a deep learning pipeline, referred to as {\ours}, which automatically estimates the remaining surgery duration (RSD) intraoperatively by using only visual information from laparoscopic videos. Previous state-of-the-art approaches for RSD prediction are dependent on manual annotation, whose generation requires expensive expert knowledge and is time-consuming, especially considering the numerous types of surgeries performed in a hospital and the large number of laparoscopic videos available. A crucial feature of {\ours} is that it does not depend on any manual annotation during training, making it easily scalable to many kinds of surgeries. The generalizability of our approach is demonstrated by testing the pipeline on two large datasets containing different types of surgeries: 120 cholecystectomy and 170 gastric bypass videos. The experimental results also show that the proposed network significantly outperforms a traditional method of estimating RSD without utilizing manual annotation. Further, this work provides a deeper insight into the deep learning network through visualization and interpretation of the features that are automatically learned.
\end{abstract}
\begin{IEEEkeywords}
Bypass, Cholecystectomy, Deep Learning, Laparoscopic Video, OR Planning, Remaining Surgery Duration.
\end{IEEEkeywords}

\section{Introduction}\label{sec_introduction}
\IEEEPARstart{P}atient safety is the number one priority in every department of healthcare institutions. In the surgical department, patient safety could be improved by reducing the duration of anesthesia and ventilation. In order to do this, an accurate prediction of the surgery duration is needed, for instance, to correctly estimate the amount of required anesthesia. In addition to improving patient safety, accurate surgery duration estimation also plays an important role in building an efficient OR management system. It would decrease the cost of a surgical facility by reducing both: (1) duration overestimation (which leads to underutilization of expensive resources \cite{childers2018}) and (2) duration underestimation (which causes overtime and high waiting time for patients) \cite{kayis2015hcms}.

However, it is still challenging to accurately predict surgery duration preoperatively due to multiple factors, such as diversity of patient conditions, surgeon's skills, and the variety of intraoperative circumstances. These factors are difficult, if not impossible, to be incorporated into the preoperative prediction model. For instance, it has been shown by \cite{travis2014bmj} that general surgeons underestimated surgery durations by 31 minutes in average, while anesthesiologists underestimated the durations by 35 minutes (29\% and 167\% error with respect to predicted durations, respectively). Such underestimation will lead to longer waiting time for patients, e.g., a large variation of waiting time (47$\pm$17 mins) over 157 cholecystectomy patients was observed in \cite{guedon2015hc}, while it typically requires 25 minutes for patient preparation.

To alleviate this problem, an adaptive scheduling which can be dynamically updated as the day progresses is often utilized. To do so, verbal communication with the surgical staff is typically used to obtain an estimate of the remaining surgery duration (RSD). However, such interruption is undesirable since it disrupts the smoothness of surgical workflow and may compromise the safety in the OR \cite{wiegmann2007}. Therefore, in this paper, we propose an \textit{automatic} way to \textit{intraoperatively} estimate the RSD. Such an automatic method has the added benefit that all required personnel can be immediately notified. Here, we focus on performing the task on laparoscopic procedures using visual information coming from the endoscope. 

It is important to note that RSD estimation using solely visual information is a challenging problem due to the fact that frames from different videos, whose visual appearance may vary significantly, may have the same remaining duration label. Here, we argue that the video frames contain discriminative characteristics which can be used to perform the RSD estimation and these characteristics can be automatically identified by using a deep learning pipeline.

In our previous work \cite{aksamentov2017miccai}, we proposed a deep learning pipeline, consisting of a convolutional neural network (CNN) and a long-short term memory (LSTM) network, to perform the RSD estimation. The CNN is trained to perform surgical phase recognition so that the network extracts semantically meaningful visual features from the images, while the LSTM is trained to perform the RSD estimation via regression. Subsequently, phase labels, which are obtained from manual annotation, are required to train the CNN. In this paper, we propose to eliminate the need for any manual annotation from the training process. This is achieved by training the CNN to perform \textit{progress estimation} instead of surgical phase recognition. Progress estimation is the task of predicting how long the surgery has progressed with respect to its expected duration. We denote the frame-wise progress labels in real values; 0 indicating the beginning of surgery and 1 as the end of surgery. Similarly to the RSD estimation, we formulate progress estimation as a regression task. Here, we train the LSTM network to perform both progress and RSD estimation in a multi-task manner. Since both progress and RSD labels can be automatically generated from the dataset, the proposed deep learning pipeline, called \ours, consequently does not require any labels from manual annotation. Therefore, {\ours} can be easily generalized to other types of surgeries and in this paper we show that it can be used to predict the RSD of surgeries much longer than cholecystectomy, i.e., gastric bypass. Not relying on manual annotation also enables all available laparoscopic videos to be used for training. This could potentially lead to better generalization across surgeries with varying patient conditions and surgeon styles. In addition to presenting the results of extensive comparisons, we also provide deeper analysis of the results, for instance by visualizing the LSTM cell values to interpret the features learned by the pipeline.

In summary, the contributions of this paper are three-fold: (1) we propose a deep learning pipeline to estimate RSD which does not require any manual annotation for the training process; (2) we show that the method is generalizable to another surgery type (i.e., bypass), which illustrates its potential to be used on other surgery types; and (3) we present the results of extensive comparisons with other approaches as well as interpretation of what the deep learning pipeline automatically learns to perform the RSD estimation.

\section{Related Work}\label{section_relatedWork}
Despite the interests in performing RSD estimation, there has only been a limited number of work addressing the task in the computer-assisted intervention community. Most early studies focus on predicting the surgery duration preoperatively, for example the Last 5 Case method \cite{macario1999} predicts the surgery duration based on the procedure-surgeon historical data. Other preoperative methods utilizing patient's age \cite{ammori2001}, operational (e.g., OR assignment and assigned surgical team) and temporal (e.g., the weekday, month, year and time of day) information \cite{kayis2012amia} have also been investigated to predict the surgery duration. However, it is still difficult for these preoperative approaches to correctly predict the surgery duration due to the uniqueness and unpredictability of each surgical procedure. 

In the literature, several studies have addressed the RSD prediction task intraoperatively. For instance, a semi-automatic method requiring the input from anesthesiologists during the surgery is presented in \cite{dexter2009}. Other signals, such as surgical tool usage \cite{padoy2008,maktabi2017} and low-level task representations (i.e., tool, organ, and action) \cite{franke2013} have also been used to perform RSD estimation. However, a semi-automatic method is disadvantageous since it disrupts the surgical workflow, and in the other studies, the signals are typically obtained through manual annotation, rendering the methods impractical for intraoperative applications. In \cite{guedon2015hc}, a classification approach using the activation of the electrosurgical devices was proposed to answer: should the next patient be called? However, the pipeline is constrained to start the detection after the procedure has progressed for 15 minutes and assumes that the next patient should be prepared 25 minutes before the surgery ends. In contrast to these studies, we propose a method that: (1) is fully automatic and does not require any human intervention; (2) uses the visual information from laparoscopic videos; and (3) continuously predicts the RSD during the procedure. 

In the computer vision community, only a few studies address the problem of estimating the remaining duration or the progress of an activity. For instance, a recent work \cite{becattini2017arxiv} proposed a deep architecture to localize short activities, such as cliff diving and tennis swing, as well as to predict its progress. In \cite{li2017arxiv}, a deep architecture was proposed to estimate progress and identify phases from various datasets, then the remaining duration is subsequently derived from the progress (as shown in Eq. \ref{eq:derived-rsd}). This work differs from these studies in three-fold: (1) our proposed pipeline does not require any labels obtained from manual annotation; (2) instead of deriving the RSD from progress, it is directly estimated via regression by the deep learning pipeline and such a direct estimation is shown to be better in our experimental results; and (3) the RSD prediction is performed on datasets consisting of long duration sequences with a high duration variance (e.g, within the dataset of cholecystectomy surgeries, the longest laparoscopic video is 8 times longer than the shortest one).

\section{Methodology}\label{section_method}

\begin{figure}
\begin{centering}
\includegraphics[width=8cm]{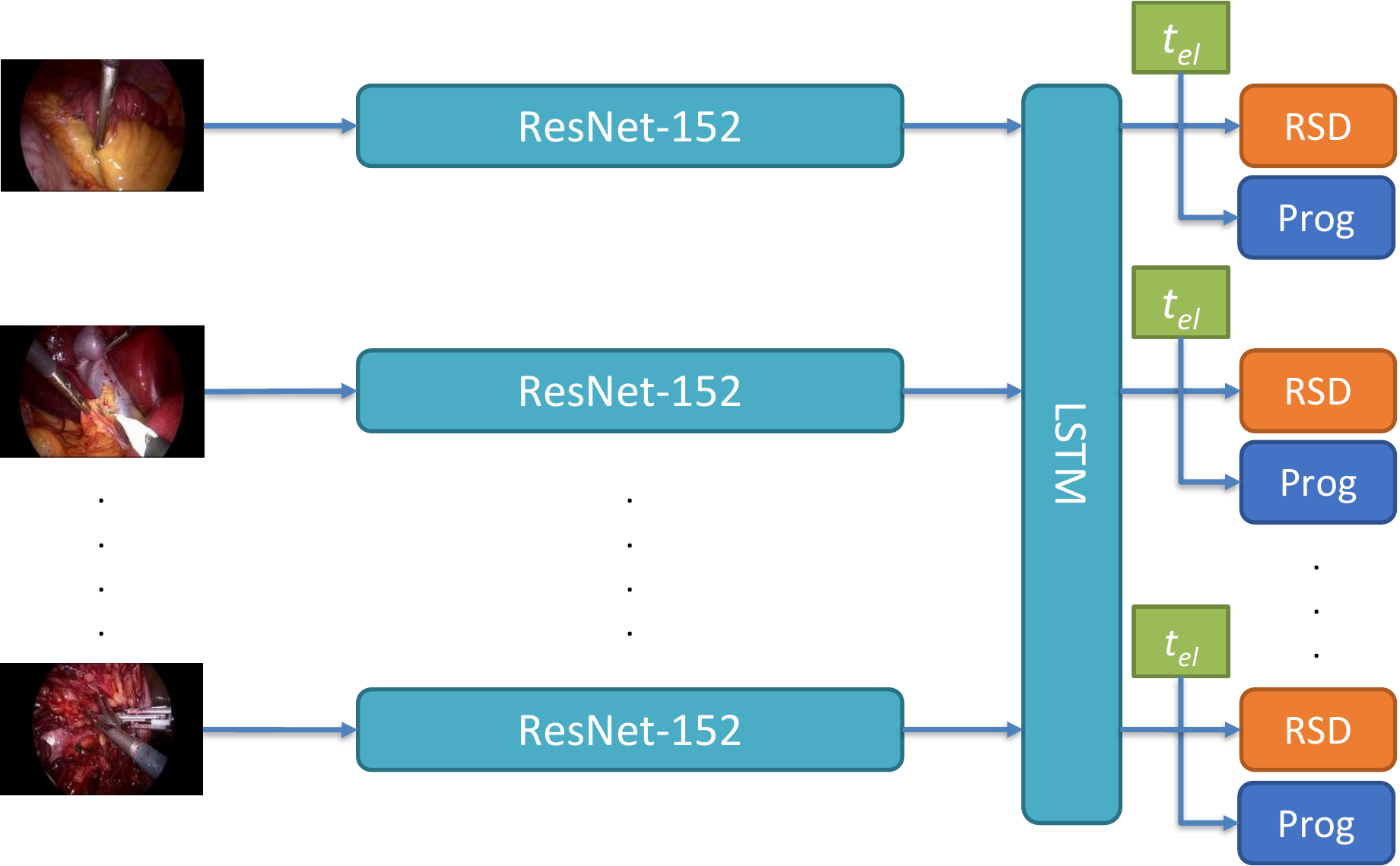}
\par\end{centering}
\caption{Architecture of \ours. \label{fig:architecture}}
\end{figure}

{\ours} consists of two elements: a convolutional neural network (CNN) and a long-short term memory (LSTM) network. The CNN is used to extract discriminative visual features from the video frames, while the LSTM network is used to incorporate the temporal information to the prediction process. The illustration of the architecture is shown in Fig. \ref{fig:architecture}.

\subsection{Deep Architecture}

Here, a deep residual network (ResNet-152) \cite{he2016deep} is chosen as the CNN architecture because such a residual network is the state-of-the-art in the computer vision community. In addition, our early experiments show that it outperforms other networks like AlexNet in tasks like surgical phase recognition (72.8\% vs. 82.3\% accuracy). We finetune the network to perform progress estimation, which is a regression task of estimating how much the surgery (in percentage) has progressed, by replacing the last layer of the ResNet-152 architecture with a fully connected layer containing a single node for regressing progress values. Sigmoid nonlinearity is applied at the output of this layer to ensure that progress values lie in the $[0,1]$ range. At training time, the video frames along with their progress label are fed to the network. As for the LSTM network, it is designed to perform two tasks in a joint manner: progress and RSD estimation. To do so, the visual features extracted using ResNet are taken as input by an LSTM layer. Before passing the output of LSTM to estimate the RSD, it is first concatenated with the elapsed time (in minute). We incorporate the elapsed time into the final feature since RSD ($t_{rsd}$), elapsed time ($t_{el}$) and progress ($prog$) have a relationship which can be expressed as:
\begin{equation}
t_{rsd} = T-t_{el} = \frac{t_{el}}{prog}-t_{el},
\label{eq:derived-rsd}
\end{equation}
where $T$ is the total duration of the surgery and $prog$ is the progress value. Therefore, we argue that the elapsed time is an essential information to be incorporated into the prediction process. The concatenated feature is then passed to two independent fully connected layers, each containing 1 node, which regress the RSD and progress prediction values, respectively. Sigmoid nonlinearity is again applied at the output of the layer regressing progress values. For both tasks, smooth L1 loss \cite{girshick2015iccv} is used and the final loss is the summation of both losses with equal weights.
 
Note that both labels, i.e., progress and RSD labels, are automatically generated from the sequences. Therefore, {\ours} does not require any manual annotation. 

\subsection{Training Strategy}

The pipeline is trained and tested at 1 fps. A 152-layer bottleneck-based ResNet model, pretrained on the ImageNet dataset, is finetuned with a batch size of 48 on our dataset; while the LSTM is trained on complete sequences.

\miniTitle{Two-step optimization.} Because ResNet is a large network and the cholecystectomy videos are of long durations, it is difficult to train the complete pipeline in an end-to-end manner due to memory constraints. To alleviate this problem, we use a two-step optimization process. First, we train the CNN by finetuning a pre-trained ResNet model. Once ResNet is finetuned, the CNN is used to extract the visual features (i.e., the second last layer of ResNet). These features are then directly passed to the LSTM network, which is trained on complete sequences.

\miniTitle{RSD Normalization.} The RSD estimation task is a regression problem with a high range target value (i.e., 0-100 min, Figure \ref{fig:cholec-dataset-distribution}, for cholecystectomy and 0-208 min, Figure \ref{fig:bypass-dataset-distribution}, for bypass). Such a high target value can only be regressed when a small regularization factor is applied on the layer weights. However, we observed that such low regularization factors always lead to overfitting. To mitigate this issue, at training time, we normalize the target values by dividing the RSD with a scalar ($s_{norm}$). At testing time, the values regressed by the pipeline are denormalized to obtain the final RSD.

\miniTitle{Hyperparameter search.} To obtain the best setup for both CNN and LSTM training, we have performed an extensive hyperparameter search on various parameters, including RSD normalization scale $s_{norm}$, regularization factor, weight decay, learning and dropout rates, and the number of LSTM states, using the training and validation subsets.

Based on the results of the hyperparameter search, we apply dropout with probability $0.3$ to the feature vector extracted from the CNN, before providing it as input to the LSTM, as well as to the output feature vector of the LSTM.

We also found that it is best to utilize a RSD normalization factor which restricts the RSD target values within the 0 to 20 range. Accordingly, we set $s_{norm}=5$ for Cholec120 and $s_{norm}=10$ for Bypass170.

CNN and LSTM models are both trained using stochastic gradient descent optimization with a momentum of $0.9$. During CNN finetuning for progress regression, the total training iterations are $50$k and an initial learning rate of $10^{-3}$ is selected. The learning rate is reduced by a factor of 10 after every $20$k iterations. The weight decay parameter is set to $5\cdot10^{-4}$.

The LSTM network is trained for a total of $30$k iterations on the Cholec120 dataset and $50$k iterations on the Bypass170 dataset, with an initial learning rate of $10^{-3}$. After every $10$k iterations on the Cholec120 dataset and $5$k iterations on the Bypass170 dataset the learning rate is decayed by a factor of $10$. A weight decay parameter of $10^{-2}$ is utilized.

\section{Experimental Setup}
 
\subsection{Cholec120 Dataset}

\begin{figure}
\begin{centering}
\includegraphics[height=4cm]{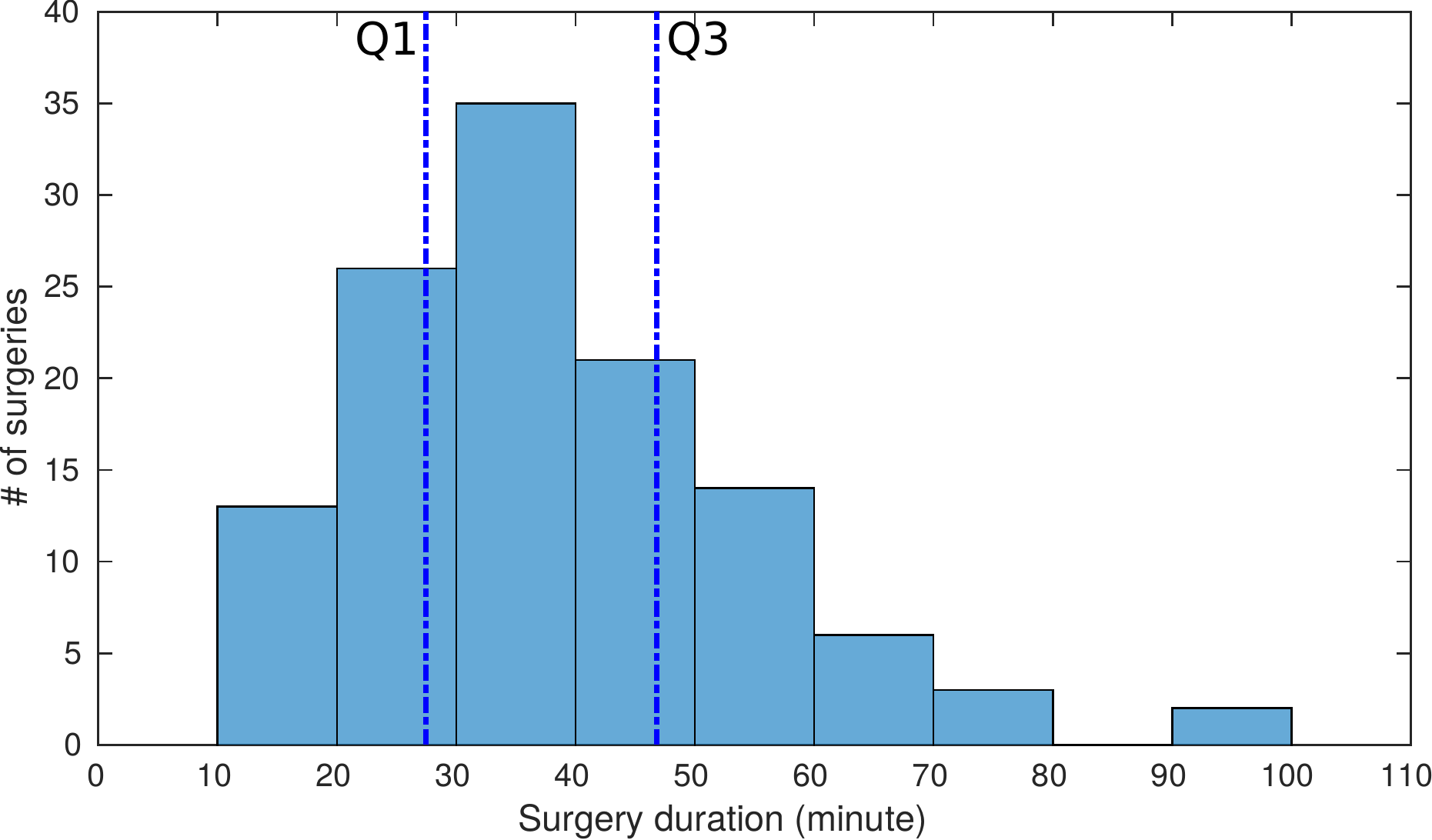}
\par\end{centering}
\caption{Distribution of the surgery duration $T$ in the Cholec120 dataset with dashed blue lines (first and third quartiles, Q1=27.4 mins and Q3=46.8 mins) indicating the boundaries of short, medium, and long surgeries. \label{fig:cholec-dataset-distribution}}
\end{figure}

The first dataset we use to evaluate the approach is \textit{Cholec120} \cite{aksamentov2017miccai}. This dataset contains 120 recordings of cholecystectomy procedures performed by 33 surgeons. This dataset is generated by combining the Cholec80 dataset \cite{twinanda2017endonet} and an additional 40 cholecystectomy videos. All 120 videos are annotated with the surgical phases defined in \cite{twinanda2017endonet}. The videos are recorded at 25 fps and accumulate over 75 hours of recordings. The videos have an average duration of 38.1 mins ($\pm$ 16.0 mins). We retain the same 4-fold cross validation setup to the one in \cite{aksamentov2017miccai} for comparison purposes. In Fig. \ref{fig:cholec-dataset-distribution}, we show the duration distribution of the dataset.

To train and test the approach, the dataset is split into 4 parts: T1 (40 videos), T2 (40 videos), V (10 videos), and E (30 videos). Subset T1 is used to train the CNN, while the combination of T1 and T2 is used to train the LSTM. The CNN is only trained on T1 to avoid overfitting on LSTM. Subset V is used as validation during both CNN and LSTM training. Subset E is used to evaluate the trained CNN-LSTM pipeline. We perform the evaluation on the dataset using a four-fold cross validation so that all videos in the dataset have been used in evaluation and we ensure that each subset has a duration distribution similar to the complete dataset. 

\subsection{Bypass170 Dataset}

\begin{figure}
\begin{centering}
\includegraphics[height=4cm]{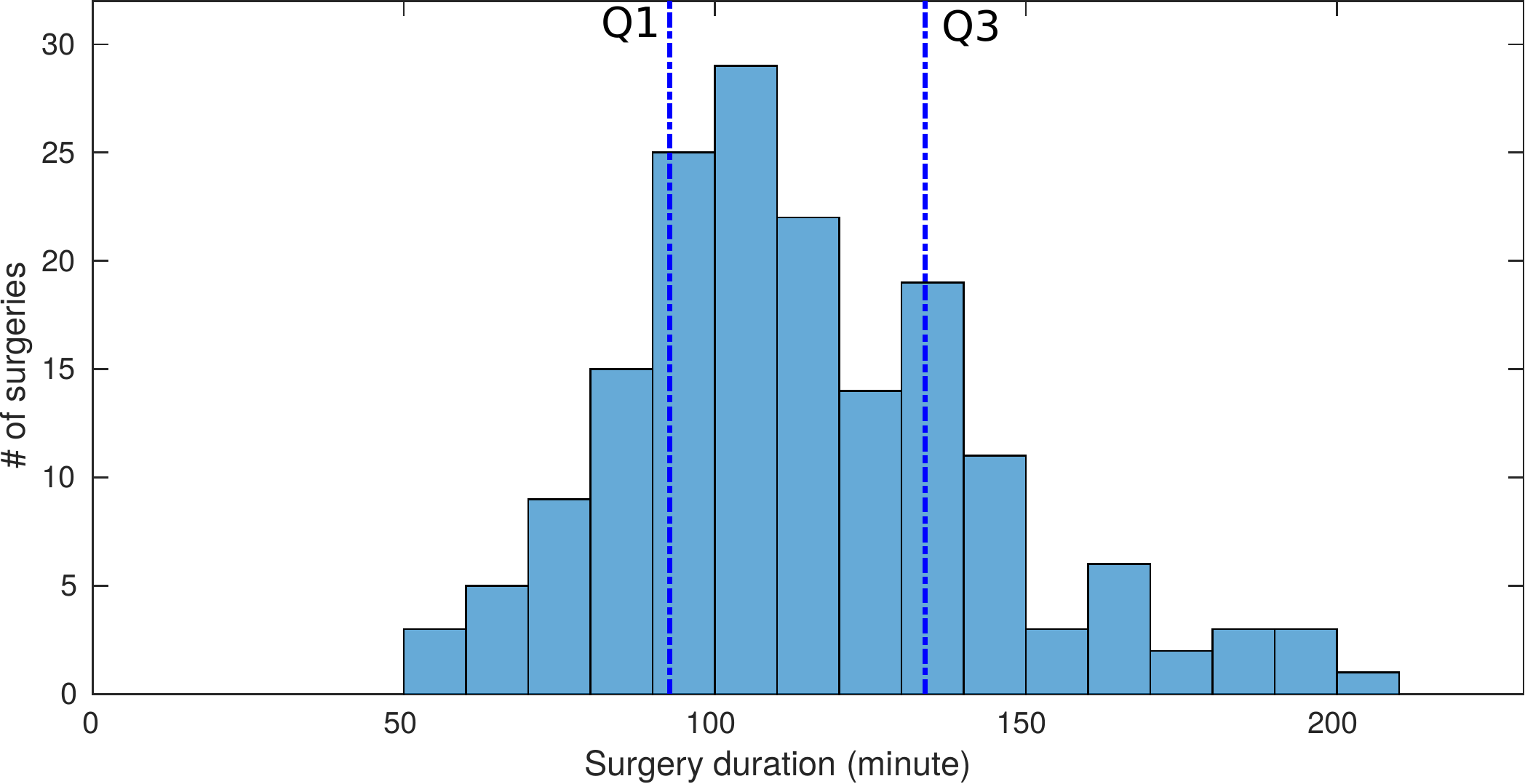}
\par\end{centering}
\caption{Distribution of the surgery duration $T$ in the Bypass170 dataset with dashed blue lines (first and third quartiles, Q1=92.7 mins and Q3=133.8 mins) indicating the boundaries of short, medium, and long surgeries. \label{fig:bypass-dataset-distribution}}
\end{figure}

To test the generalizability of our proposed approach, we collect a large dataset of 170 bypass videos containing over 327 hours of recordings, called \textit{Bypass170}. The procedures were performed by a total of 28 surgeons. Thanks to the fact that our approach only requires automatically generated labels, we did not need to manually label this large dataset to perform the RSD prediction. We chose to evaluate the robustness of the approach on bypass surgeries since they are frequently performed at the University Hospital of Strasbourg and as they are on average three times longer than the cholecystectomy procedures: the average duration of this dataset is 115 ($\pm$ 29) mins. This difference can also be observed from the dataset distribution shown in Fig. \ref{fig:bypass-dataset-distribution}.

For evaluation, we use a similar scheme by splitting the dataset into four subsets (T1, T2, V, and E subsets). Instead of performing the evaluation in a cross validation manner, we took advantage of the large number of videos in this dataset to investigate which part of the two-step optimization is more important: the CNN or the LSTM side. To do so, with fixed subsets T1+T2, V and E (120, 10 and 40 videos, respectively), we varied the size of T1, i.e., the videos used for CNN training, as either 40, 60 or 80 videos. All training videos, T1+T2, were again used to train the LSTM.

\begin{table*}[h]
\begin{centering}
\begin{tabular}{|l|c|c|c|c||c|c|c|c|}
\hline 
\multicolumn{2}{|c|}{\multirow{2}{*}{Method}}  & \multirow{2}{*}{CNN} & \multirow{2}{*}{LSTM}& \multirow{1}{*}{Manual} & \multicolumn{4}{c|}{MAE in minute}\tabularnewline
\cline{6-9} 
\multicolumn{2}{|c|}{} & & & Annotation & Complete  & Short  & Medium  & Long \tabularnewline
\hline 
\hline 
\multirow{2}{*}{Na\"ive}  & Mean & n/a & n/a& No& 11.1$\pm$8.0 & 17.3$\pm$3.9 & 5.0$\pm$3.0 & 17.3$\pm$8.7 \tabularnewline
\cline{2-9} 
 & Median & n/a & n/a& No & 10.7$\pm$8.0 & 14.3$\pm$3.8 & 4.7$\pm$2.5 & 19.1$\pm$8.5 \tabularnewline
\hline 
\multirow{2}{*}{PhaseInferred-GT}  & Mean & n/a & n/a & Yes& 8.1$\pm$5.8 & 10.9$\pm$3.4 & \textbf{4.1$\pm$1.8} & 13.5$\pm$6.6 \tabularnewline
\cline{2-9} 
 & Median & n/a & n/a& Yes & 8.0$\pm$6.6 & \textbf{6.6$\pm$2.9} & 4.4$\pm$2.7 & 16.9$\pm$6.6 \tabularnewline
\hline 
\hline
\multirow{2}{*}{PhaseInferred-LSTM}  & Mean & Phase & Phase & Yes& 8.5$\pm$6.2 & 11.7$\pm$3.8 & 4.3$\pm$1.6 & 14.1$\pm$7.3 \tabularnewline
\cline{2-9} 
 & Median & Phase & Phase& Yes & 8.4$\pm$6.8 & 7.3$\pm$3.1 & 4.5$\pm$2.6 & 17.5$\pm$7.1 \tabularnewline
\hline 
\multicolumn{2}{|l|}{TimeLSTM \cite{aksamentov2017miccai}} & Phase & RSD & Yes& \textbf{7.7$\pm$5.2} & 9.9$\pm$3.9 & 4.8$\pm$2.2 & \textbf{11.2$\pm$7.0} \tabularnewline
\hline
\hline
\multicolumn{2}{|l|}{Single-task} & Progress & RSD& No & 8.5$\pm$5.2 & 9.2$\pm$3.8 & 6.5$\pm$3.3 & 12.1$\pm$6.9
  \tabularnewline
\hline 
\multicolumn{2}{|l|}{{\ours} (Progress-Derived)} & Progress & RSD-Progress& No & 9.6$\pm$4.7 & 7.3$\pm$3.2 & 8.7$\pm$3.1 & 13.7$\pm$5.9 \tabularnewline
\hline 
\multicolumn{2}{|l|}{\ours} & Progress & RSD-Progress& No & 8.1$\pm$5.4 & 7.5$\pm$4.2 & 5.7$\pm$2.6 & 13.4$\pm$6.9 \tabularnewline
\hline
\end{tabular}\tabularnewline
\par\end{centering}
\vspace{0.2cm}
\caption{RSD prediction results on Cholec120. The MAEs (mean$\pm$std) are shown for the complete dataset as well as for short, medium, and long surgeries. The best MAEs for each column are shown in bold. Additionally the CNN and LSTM columns mention the task used to train the specific networks. \label{tab:cholec-rsd-results}}
\end{table*}

\begin{table*}[h]
\begin{centering}
\begin{tabular}{|l|c|c|c||c|c|c|c|}
\hline 
\multicolumn{2}{|c|}{\multirow{2}{*}{Method}}  & \multirow{2}{*}{CNN} & \multirow{2}{*}{LSTM} & \multicolumn{4}{c|}{MAE in minute}\tabularnewline
\cline{5-8} 
\multicolumn{2}{|c|}{} & & &  Complete  & Short  & Medium  & Long \tabularnewline
\hline 
\hline
\multirow{2}{*}{Na\"ive}  & Mean & n/a & n/a & 22.5$\pm$15.6 & 32.8$\pm$8.9 & 11.1$\pm$6.8 & 35.1$\pm$17.2  \tabularnewline
\cline{2-8}
 & Median & n/a & n/a & 21.3$\pm$16.5 & 27.9$\pm$8.9 & \textbf{9.4$\pm$7.6} & 38.7$\pm$16.7 \tabularnewline
\hline 
\hline
\multirow{3}{*}{\ours} & 40-120 & \multirow{3}{*}{Progress} & \multirow{3}{*}{RSD-Progress} & \textbf{15.6$\pm$7.9}& 15.8$\pm$5.7& 11.6$\pm$4.2& 23.2$\pm$9.6 \tabularnewline
\cline{2-2}\cline{5-8} 
& 60-120 & & &15.8$\pm$7.5& \textbf{15.2$\pm$4.7}& 12.5$\pm$4.6& \textbf{23.1$\pm$9.4}\tabularnewline
\cline{2-2}\cline{5-8}
& 80-120 & & & 17.0$\pm$8.1 & 16.7$\pm$3.1 & 13.7$\pm$5.6 & 24.0$\pm$11.5
\tabularnewline
\hline
\end{tabular}\tabularnewline
\par\end{centering}
\vspace{0.2cm}
\caption{RSD prediction results on Bypass170. The MAEs (mean$\pm$std) are shown for the complete dataset as well as for short, medium, and long surgeries. The best MAEs for each column are shown in bold. Additionally the CNN and LSTM columns mention the task used to train the specific networks.\label{tab:bypass-rsd-results}}
\end{table*}

\subsection{Baselines}

We compare our proposed approach with the following baselines:
\begin{itemize}
\item \textbf{Na\"{i}ve approach}\cite{aksamentov2017miccai} \textbf{-} This approach uses the historical data of the surgeries. The approach is as follows: at time $t_{el}$ during a surgery, the RSD $t_{rsd}$ is obtained by computing $max(0,t_{ref}-t_{el})$, where $t_{ref}$ is a referencial duration derived from the dataset (e.g., mean or median). The $max(\cdot,\cdot)$ operator is used to ensure that $t_{rsd}$ is always positive. As can be seen, this method does not require any manual annotation, which is beneficial. However, it also does not take into account any intraoperative information and only relies on the statistics of the historical data. 

\item \textbf{Phase-inferred approach} \cite{aksamentov2017miccai} \textbf{-} This approach incorporates intraoperative information by leveraging from the fact that the execution of a surgery is governed by a surgical workflow which dictates the sequentiality of the phases during the procedure. If the current phase is known, one can then estimate the RSD in a way that is finer than the Na\"ive approach. The RSD $t_{rsd}$ is computed as $max(0,t^{p}_{ref}-t^{p}_{el})+\sum^{N}_{m=p+1}{t^m_{ref}}$ where $t^m_{ref}$ is either the mean or median duration of phase $m$, $t^{p}_{el}$ is the elapsed time in current phase $p$, and $N=7$ is the number of defined phases. In this experiment, the phase information is obtained in two different manners: (1) ground-truth (GT) obtained from manual annotation and (2) predicted phase from a CNN-LSTM network trained to perform phase recognition. Having phase information from manual annotation is equivalent to having an expert inside the OR marking the transitions between phases during the surgery, while the deep network detects the phase transitions automatically. Such a deep network would still require manual annotations during training.

\begin{sloppypar}
Due to the fact that the distribution of the surgery durations is asymmetric (as shown in Fig. \ref{fig:cholec-dataset-distribution} and \ref{fig:bypass-dataset-distribution}), where long cases (outliers) could inflate the average estimated case duration, we choose to use both mean and median of surgery durations for $t_{ref}$ and $t_{ref}^{p}$. These referential durations are obtained using the surgeries included in the training set (T1+T2).
\end{sloppypar}

\item \textbf{TimeLSTM} \cite{aksamentov2017miccai} \textbf{-} Similarly to our proposed approach, this deep learning pipeline consists of ResNet and LSTM. However, here ResNet is first finetuned to perform phase recognition \cite{twinanda2017endonet} before being used to extract the visual features. Then, the LSTM network is trained to perform RSD prediction. Note that this pipeline requires manually obtained labels to train the CNN part of the pipeline.

\item \textbf{Single-task LSTM -} In this pipeline, we train the CNN to perform progress estimation and the LSTM is trained to solely perform RSD estimation. This is done by removing the progress estimation task from the LSTM training in {\ours} (see Fig. \ref{fig:architecture}). We do so in order to show that it is beneficial to design a multi-task LSTM to perform RSD estimation.

\item \textbf{{\ours} progress-derived approach} exploits the fact that the RSD can be derived from the progress estimation using Eq. \ref{eq:derived-rsd}. This approach has been utilized to estimate the remaining duration of activities in \cite{li2017arxiv}. We compare to this approach in order to investigate whether the derived RSD is better than the direct RSD estimation. 
\end{itemize}

\subsection{Evaluation Metrics}

We use mean absolute error (MAE in minute) for RSD estimation to measure the performance of the methods. This metric is used since it is the natural metric to be used for such a regression problem, especially for RSD estimation since it is easily interpretable by clinicians and enables comparison of different methods in a standard manner.

In addition to showing this metric on the complete dataset, we are also comparing the methods' performances on short, medium, and long surgeries, which respectively contain 25\%, 50\%, and 25\% of the dataset. The short surgeries are videos which fall on the left side of Q1 ($T<$ Q1); the long surgeries are videos on the right side of Q3 ($T>$ Q3), as depicted in Fig. \ref{fig:cholec-dataset-distribution} and \ref{fig:bypass-dataset-distribution}. The rest of the videos are categorized as medium surgeries. We look at the performance on different ranges in order to observe the effectiveness of {\ours} in modeling variation in the surgical workflow.

We also present discussions beyond quantitative analysis in Section \ref{sec:discussion} where we deliberate several interesting points, such as the effect of CNN finetuning, the reliability of the approach, and deeper analysis of the LSTM network.

\section{Experimental Results \label{sec:results}}

\subsection{Cholec120 Dataset}

In Table \ref{tab:cholec-rsd-results}, we show the MAE of all methods on the complete Cholec120 dataset as well as on short, medium, and long surgeries. We can see that on the complete dataset, using the median as the referential duration ($t_{ref}$) results in better performance than using the mean. This is due to the skewed nature of the dataset distribution. For such a distribution, the median will be less sensitive to outliers than the mean. It is also shown that the na\"ive approach yields the worst result, being significantly outperformed by RSDNet ($p < 0.005$)\footnotemark[1]. This is expected since the approach does not take into account any intraoperative information. 

By incorporating the phase information into the RSD estimation process, the phase-inferred approaches\footnotemark[2] show significant improvements compared to the na\"ive approach ($p<0.005$). It can be seen that the semi-automatic (phase labels from GT) and the automatic (phase detection from LSTM) methods yield similar results. In other words, we could remove the expert observer in the RSD estimation process without sacrificing the performance of the system, thanks to the high performance of the phase recognition pipeline, yielding 87\% of accuracy \cite{yengera2018}.

\addtocounter{footnote}{0}
\stepcounter{footnote}\footnotetext{All $p$-values in this paper are computed using the one-sided t-test, using the MAE as a measure of model performance.}
\stepcounter{footnote}\footnotetext{Due to a miscalculation in the lengths of one surgical phase in \cite{aksamentov2017miccai}, the results of the phase-inferred approaches presented in Table \ref{tab:cholec-rsd-results} have been recalculated.}

Additionally, we observe that the direct RSD estimation of RSDNet is significantly better than the approach of indirectly obtaining RSD through progress estimates using Eq. \ref{eq:derived-rsd} as done in prior work \cite{li2017arxiv} ($p<0.05$). Also, the utilization of progress information during training does help improve the RSD estimation performance, making RSDNet perform slightly better than the single-task network.

Ultimately, we can observe that {\ours} yields similar performance as compared to TimeLSTM and the phase-inferred approaches. This shows that either phase labels or progress labels can be effectively utilized for predicting RSD. In practice, it is more advantageous to use our proposed RSDNet, which relies on the progress labels, since these labels are automatically-generated. This dramatically reduces the human annotation effort required to train such deep networks. This also enables the architecture of {\ours} to be easily adapted to perform RSD estimation on other types of surgeries.

Observing the results on different video duration categories, we can find that the best performing method varies for every category. For medium surgeries, the methods perform more or less similar as expected, since the methods are statistics-based and the videos in the middle range contains videos close to the mean and median durations. As for other videos, methods using mean as referential duration tend to favor long videos, while methods using median favor short videos. This is due to the fact that the value of median is lower than mean. However, a satisfactory approach should be robust to the surgery duration and accurate across different categories. It can be seen in Table \ref{tab:cholec-rsd-results} that our proposed approach yields results that are comparable to the results of the best performing method in each category.

\subsection{Bypass170 Dataset}

In Table \ref{tab:bypass-rsd-results}, we show the performance of Na\"ive and {\ours} on Bypass170. The goal of this experiment is to show that the proposed approach generalizes to another type of surgery and outperforms a traditional method used for RSD estimation which also does not require manual annotation, i.e., the Na\"ive approach. Other methods, such as phase inferred and TimeLSTM, cannot be compared to as they require phase labels which are not available in this dataset. This also highlights the scalability of methods which do not rely on manual annotations to different kinds of surgeries. 

From the table, it can clearly be seen that RSDNet significantly outperforms the Na\"ive approach ($p<0.05$). One can see that for medium surgeries, the Na\"ive approach slightly outperforms {\ours} with a difference in MAE of around 2 minutes. This is however expected since the length of medium surgeries are very close to the referential duration ($t_{ref}$), thus the Na\"ive approach is practically handcrafted to predict surgeries with medium duration. This behavior is also observed on Cholec120 (see Table \ref{tab:cholec-rsd-results}), but with a smaller difference in MAE of around 1 minute, since the duration of cholecystectomy surgeries are shorter and exhibit less variance than bypass surgeries. In contrast, it can be seen that {\ours} yields considerably better results for short and long surgeries with MAE difference greater than 5 and 15 minutes respectively. This shows that the pipeline is capable of modeling the variation of surgery execution. This also shows that {\ours} is generalizable to bypass surgeries and has the potential to be used to estimate RSD on other laparoscopic surgeries.

Additionally it can be observed from the table that the best result on the complete dataset was obtained by using 40 videos for training the CNN. Using 80 videos instead of 40 for CNN training showed a noticeable degradation in performance. However there was no significant difference between the results obtained using either 40 or 60 videos. In general, as it was done with Cholec120, using half the number of total training videos for CNN training seems to be effective.

\section{Discussions \label{sec:discussion} }

\subsection{RSD Prediction Analysis}
\begin{figure}[ht]
\begin{centering}
\includegraphics[width=7.8cm]{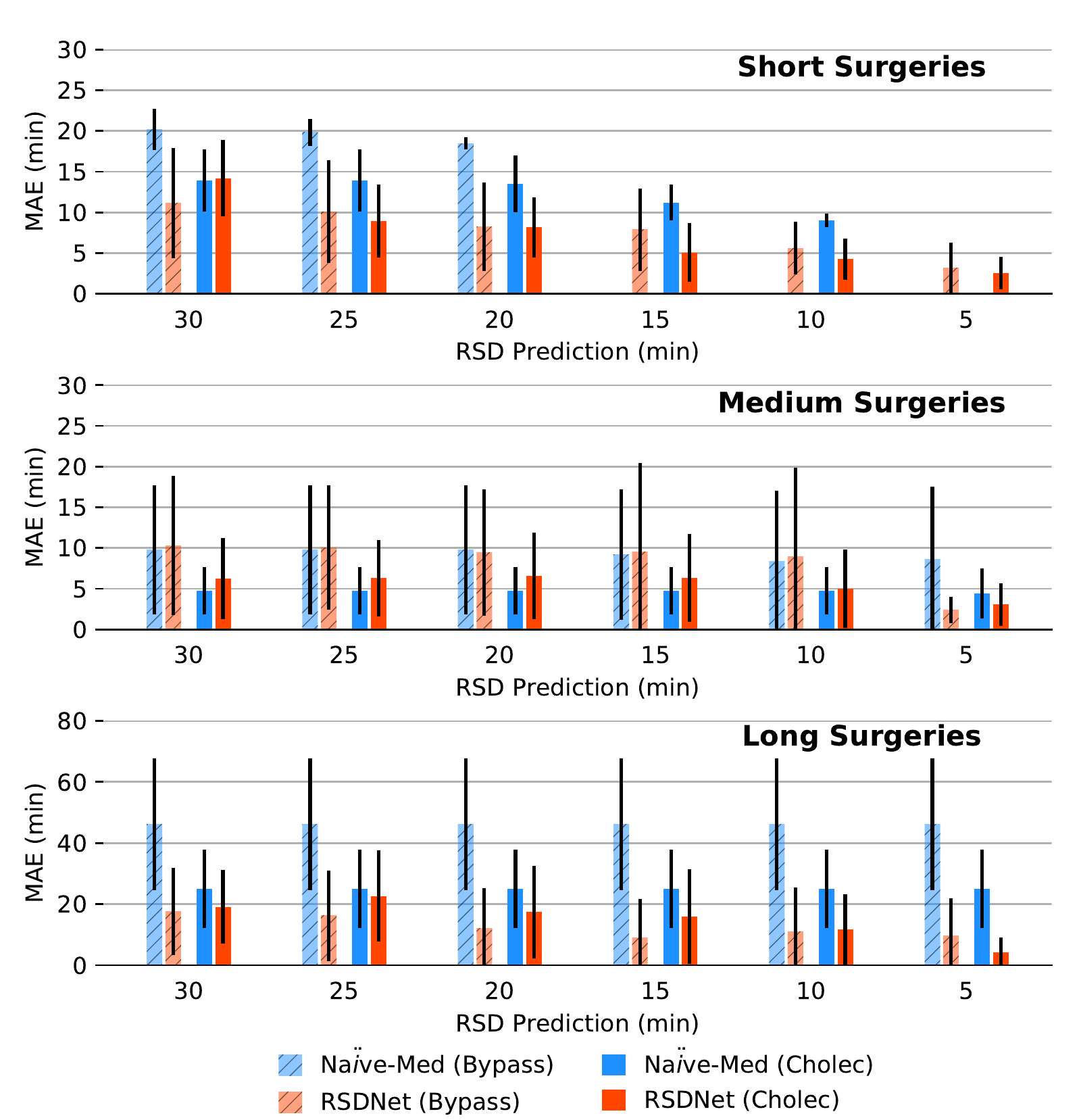}
\par\end{centering}
\caption{Graphs to show the reliability of the RSD predictions (MAE vs. RSD prediction) of RSDNet and Na\"ive approach at the end of the surgery on Cholec120 and Bypass170 datasets. (Best seen in color.)\label{fig:pred_mae}}
\end{figure}

To better understand how accurate the RSD predictions are for practical applications, we investigate the reliability of the predictions by computing the MAEs with respect to several RSD predictions (from 5 to 30 min.), shown in Fig. \ref{fig:pred_mae}. In other words, this evaluation indicates how big the error is when the method predicts that the surgery will end in, for instance, 25 minutes. A reliable RSD prediction of 25 minutes is particularly important as it is the time required for preparing the next patient for both cholecystectomy \cite{guedon2015hc} and bypass surgeries \cite{lindauer2014}. The performance of RSDNet and the Na\"ive approach are compared since these approaches do not rely on manual annotations, enabling them to be more easily scaled up to different surgery types. Comparison on both Cholec120 and Bypass170 datasets are depicted in Fig. \ref{fig:pred_mae}.

As expected from Tables \ref{tab:cholec-rsd-results} and \ref{tab:bypass-rsd-results}, the performance of RSDNet is similar to the Na\"ive approach on medium surgeries, while performing significantly better on short and long surgeries. The improvement in performance for short and long surgeries is more pronounced on the Bypass170 dataset which possesses greater variation in surgery duration (Fig. \ref{fig:bypass-dataset-distribution}). This highlights the advantage of a method like RSDNet which is able to model the progression of the surgical workflow.

Note that for all short surgeries in the Bypass170 dataset, the surgery gets completed before the Na\"ive approach makes a RSD prediction of 15 minutes or lower. Similarly for all short surgeries in the Cholec120 dataset, the surgery ends before the Na\"ive approach predicts the RSD to be 5 minutes or lower. Hence, the MAE is undefined in these cases and is not depicted in Fig. \ref{fig:pred_mae}.

\begin{figure}[h]
\begin{centering}
\includegraphics[width=7.8cm]{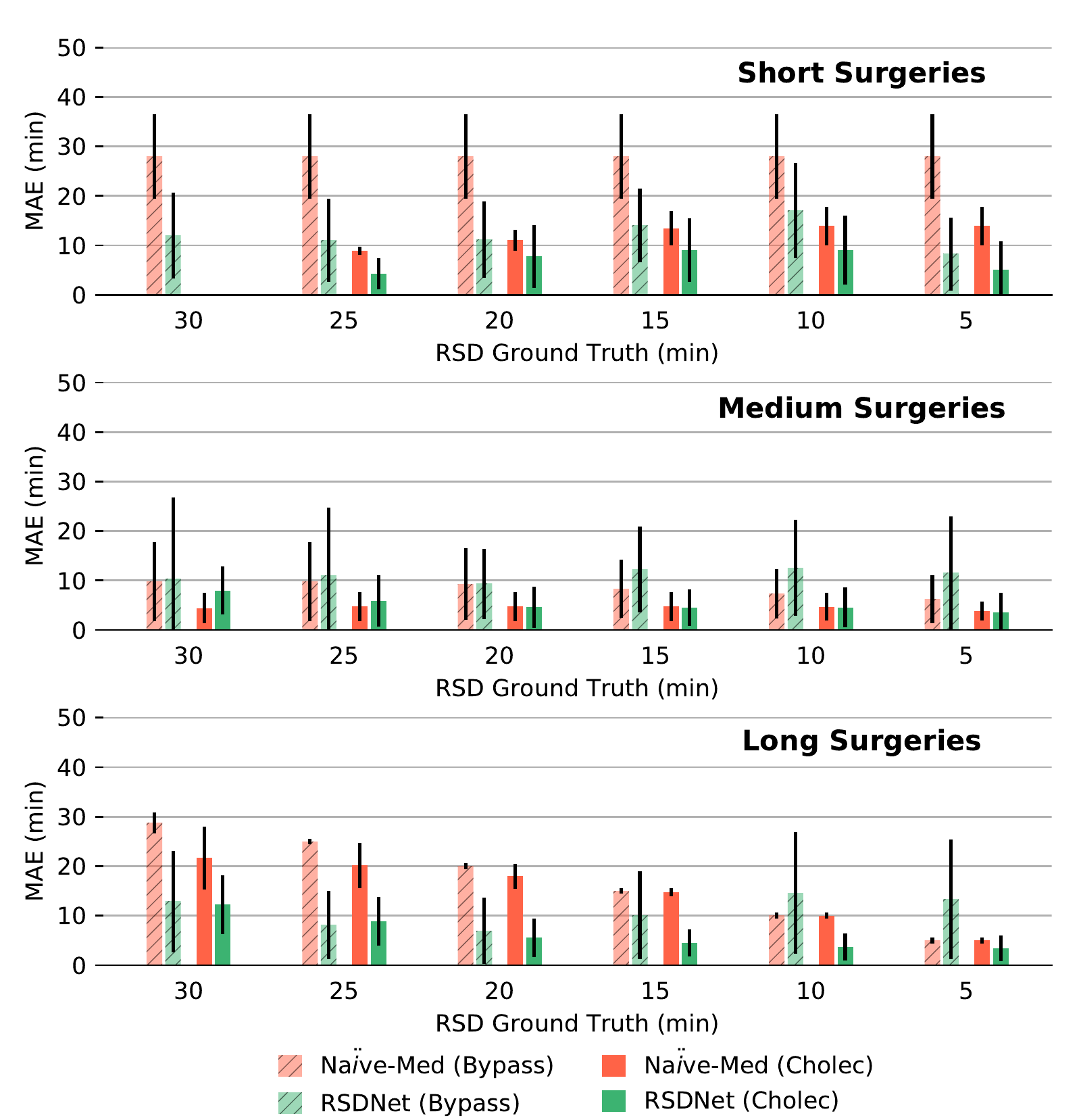}
\par\end{centering}
\caption{Graphs to show the accuracy of the RSD predictions (MAE vs. RSD ground truth) of RSDNet and Na\"ive approach at the end of the surgery on Cholec120 and Bypass170 datasets. (Best seen in color.)\label{fig:gt_mae}}
\end{figure}

In addition to comparing the reliability of RSD predictions, we compare the accuracy of RSDNet and the Na\"ive approach at specific time points in the surgery by evaluating MAEs with respect to different RSD ground truth values (5 to 30 min.), shown in Fig. \ref{fig:gt_mae}. This study provides a deeper insight into the performance of the RSD prediction models. The results for both Cholec120 and Bypass170 datasets are presented in Fig. \ref{fig:gt_mae}.

Again it can be observed that RSDNet outperforms the Na\"ive approach for short and long surgeries, while it has similar levels of performance for medium surgeries. For the Bypass170 dataset, the MAE of the Na\"ive approach towards the end of long surgeries (RSD ground truth of 10 min. and under) is lower than that of RSDNet. Since these surgeries are of considerably longer duration than the median bypass surgery duration, the Na\"ive approach would be predicting the RSD to be 0 for a significant amount of time. In fact, for all long bypass surgeries, when as much as 25 minutes are remaining in the surgery, the Na\"ive approach predicts that the surgery has been completed. This can be observed in Fig. \ref{fig:gt_mae}, where for long bypass surgeries, when the RSD ground truth is 25 minutes, the MAE of the Na\"ive approach is also 25 ($\pm0$) minutes. Hence, the small MAE values of the Na\"ive approach at the end of long bypass surgeries are not very meaningful.

It can also be seen for medium surgeries in the Bypass170 dataset, Fig. \ref{fig:gt_mae}, that the standard deviations of MAEs are greater for RSDNet as compared to the Na\"ive approach. This can be explained by the large variation in surgery duration in the Bypass170 dataset. While RSDNet models the progression of surgeries of all durations, which is particularly challenging for the Bypass170 dataset, the Na\"ive method is essentially handcrafted for medium surgeries. The slight improvement in performance for medium surgeries from the Na\"ive approach comes at the cost of considerably poorer performance on short and long surgeries.

It is also to be noted that all short surgeries in the Cholec120 dataset have a duration of less than 30 minutes. Hence, the MAE for a RSD ground truth of 30 minutes is undefined and consequently not depicted in Fig. \ref{fig:gt_mae}.

\subsection{Performance Analysis}

\begin{table*}[t]
\centering
\begin{tabular}{l c|c|c|c|c|c|c|c|c|c|c|}
\cline{3-12}
& &\multicolumn{2}{c|}{$1^{\textnormal{st}}$ Quarter} & \multicolumn{2}{c|}{$2^{\textnormal{nd}}$ Quarter} & \multicolumn{2}{c|}{$3^{\textnormal{rd}}$ Quarter} & \multicolumn{2}{c|}{$4^{\textnormal{th}}$ Quarter} & \multicolumn{2}{c|}{Full}\\
\cline{3-12}
& & Error & Fraction & Error& Fraction & Error& Fraction & Error & Fraction & Error & Fraction\\
\hline
\multicolumn{1}{|c|}{\multirow{3}{*}{Short}} & \multicolumn{1}{c|}{\multirow{1}{*}{Under-Estimation}} & -13.6$\pm$7.5 & 0.02& -5.8$\pm$3.0 & 0.27& -4.3$\pm$2.8 & 0.31& -2.2$\pm$1.5 & 0.17& -6.2$\pm$4.3 & 0.19\\
\cline{2-12}
\multicolumn{1}{|c|}{}& \multicolumn{1}{c|}{\multirow{1}{*}{Over-estimation}} & 25.0$\pm$9.3 & 0.98& 17.5$\pm$8.6 & 0.73 &10.3$\pm$5.8 & 0.69& 12.8$\pm$4.6 & 0.83& 17.6$\pm$5.0 & 0.81\\
\cline{2-12}
\multicolumn{1}{|c|}{}& \multicolumn{1}{c|}{MAE} & 24.8$\pm$9.4 & - & 16.3$\pm$8.9 & -&10.5$\pm$5.1 & -&11.7$\pm$4.8& -& 15.8$\pm$5.7& -\\ 
\hline
\multicolumn{1}{|c|}{\multirow{3}{*}{Medium}} & \multicolumn{1}{c|}{\multirow{1}{*}{Under-Estimation}} & -11.9$\pm$7.7 &0.43& -8.8$\pm$7.1&0.35& -6.6$\pm$4.8&  0.41& -2.8$\pm$1.1& 0.15&-9.0$\pm$5.7&0.34\\     
\cline{2-12}
\multicolumn{1}{|c|}{}& \multicolumn{1}{c|}{\multirow{1}{*}{Over-estimation}} &9.0$\pm$5.5 & 0.57& 10.9$\pm$5.5 &0.65& 10.6$\pm$9.8&0.59 & 11.6$\pm$5.7 & 0.85&11.4$\pm$4.4&0.66\\
\cline{2-12}
\multicolumn{1}{|c|}{}& \multicolumn{1}{c|}{MAE} & 12.5$\pm$6.7 & -& 13.0$\pm$5.5 & -& 10.3$\pm$6.3& - & 10.7$\pm$5.6& - & 11.6$\pm$4.2& - \\    
\hline\multicolumn{1}{|c|}{\multirow{3}{*}{Long}} & \multicolumn{1}{c|}{\multirow{1}{*}{Under-Estimation}} & -38.0$\pm$19.7 & 0.98 & -25.4$\pm$15.8&0.78&-17.2$\pm$8.2&0.70 &-6.5$\pm$4.4 &0.39& -25.5$\pm$11.5&0.71\\        
\cline{2-12}
\multicolumn{1}{|c|}{}& \multicolumn{1}{c|}{\multirow{1}{*}{Over-estimation}} & 3.7$\pm$1.4&0.02& 6.1$\pm$4.3& 0.22 & 10.4$\pm$7.2& 0.30 & 11.6$\pm$4.8 &0.61&11.4$\pm$4.3&0.29\\
\cline{2-12}
\multicolumn{1}{|c|}{}& \multicolumn{1}{c|}{MAE} & 37.7$\pm$20.0 & -& 25.8$\pm$15.2 & -&  18.7$\pm$6.7 & -& 10.6$\pm$5.2 & -& 23.2$\pm$9.6& -\\              
\hline\multicolumn{1}{|c|}{\multirow{3}{*}{Complete}} & \multicolumn{1}{c|}{\multirow{1}{*}{Under-Estimation}} & -18.9$\pm$16.2& 0.46&-12.9$\pm$12.8& 0.44&  -8.7$\pm$7.4 &0.46& -3.6$\pm$3.0&  0.22& -12.4$\pm$10.6&0.39\\
\cline{2-12}
\multicolumn{1}{|c|}{}& \multicolumn{1}{c|}{\multirow{1}{*}{Over-estimation}} & 12.9$\pm$10.3&0.54&12.0$\pm$7.4& 0.56& 10.5$\pm$8.4& 0.54 & 11.9$\pm$5.2& 0.78 & 13.0$\pm$5.3&0.61\\
\cline{2-12}
\multicolumn{1}{|c|}{}& \multicolumn{1}{c|}{MAE} & 21.9$\pm$15.9& - &17.0$\pm$11.0 & -&   12.5$\pm$7.1& - & 10.9$\pm$5.4& - &15.6$\pm$7.9& -\\        
\hline
\end{tabular}
\caption{Study of the amount of under-estimation and over-estimation of remaining time by the RSDNet model on the Bypass170 dataset. The values are shown for short, medium and long surgeries as well as the complete dataset. The error values and fraction of time steps where the RSD is either under-estimated or over-estimated are presented for the full surgery and for each quarter of the surgery.}\label{tab:under_over}
\end{table*}

Table \ref{tab:under_over} shows the performance and highlights the tendency of the RSDNet model to over-estimate and under-estimate RSD in different quarters of surgeries. The study is performed on the Bypass170 dataset, since bypass surgeries show high variation in surgery duration. In addition to the error values, the fraction of the surgery duration where the RSD is over-estimated or under-estimated is computed. For example, if the actual duration of a surgery is 100 seconds and the RSDNet model, making predictions at intervals of 1 second, over-estimates the RSD at 80 time-steps, then the fraction of over-estimation is 0.8. Correspondingly the fraction of under-estimation is 0.2. The values are computed for the complete dataset as well as for short, medium and long surgeries. It can be seen that short surgeries are usually over-estimated and long surgeries are under-estimated. This is expected as these surgeries deviate from the norm. In general it can be seen that the model tends to over-estimate RSD in the final quarter of surgeries. This leads us to conclude that given small values of RSD predictions, we can expect that the surgery will end within the predicted time.

\subsection{CNN Task Formulation}
 
\begin{figure*}
\centering
\begin{subfigure}[t]{0.49\textwidth}
\includegraphics[width = 8cm]
{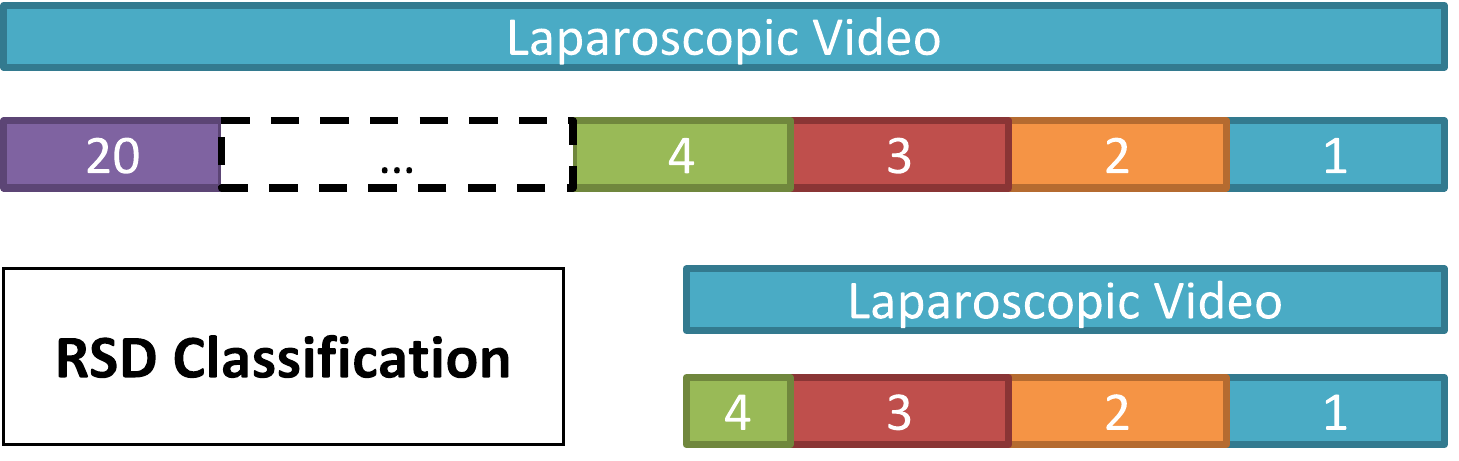}\caption{}
\end{subfigure}
\begin{subfigure}[t]{0.49\textwidth}
\includegraphics[width = 8cm]
{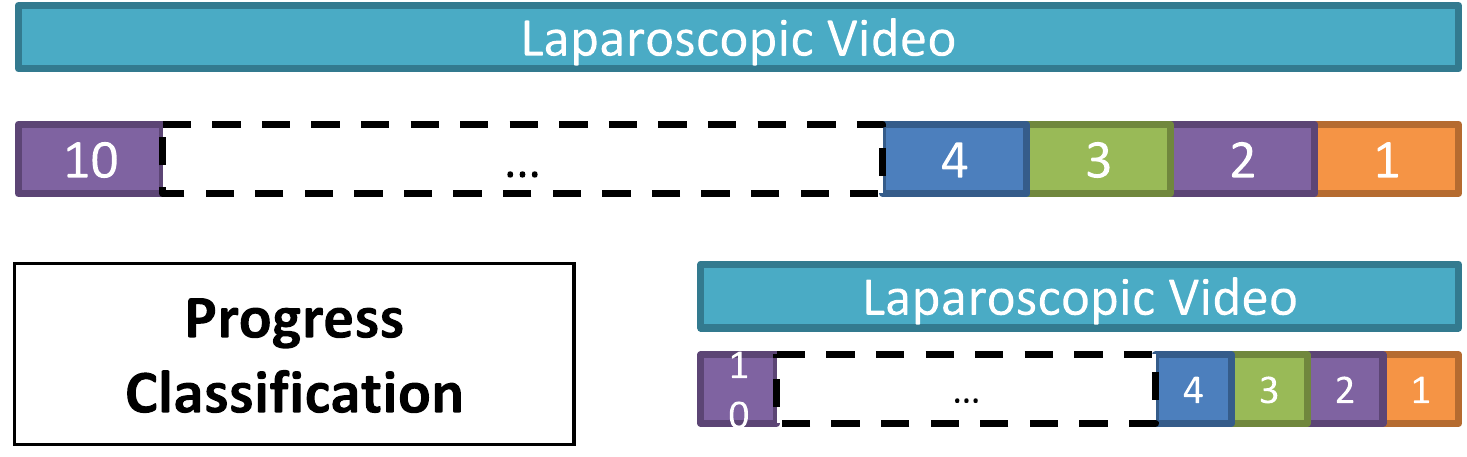}\caption{}
\end{subfigure}
\begin{subfigure}[t]{0.49\textwidth}
\includegraphics[width = 8cm]
{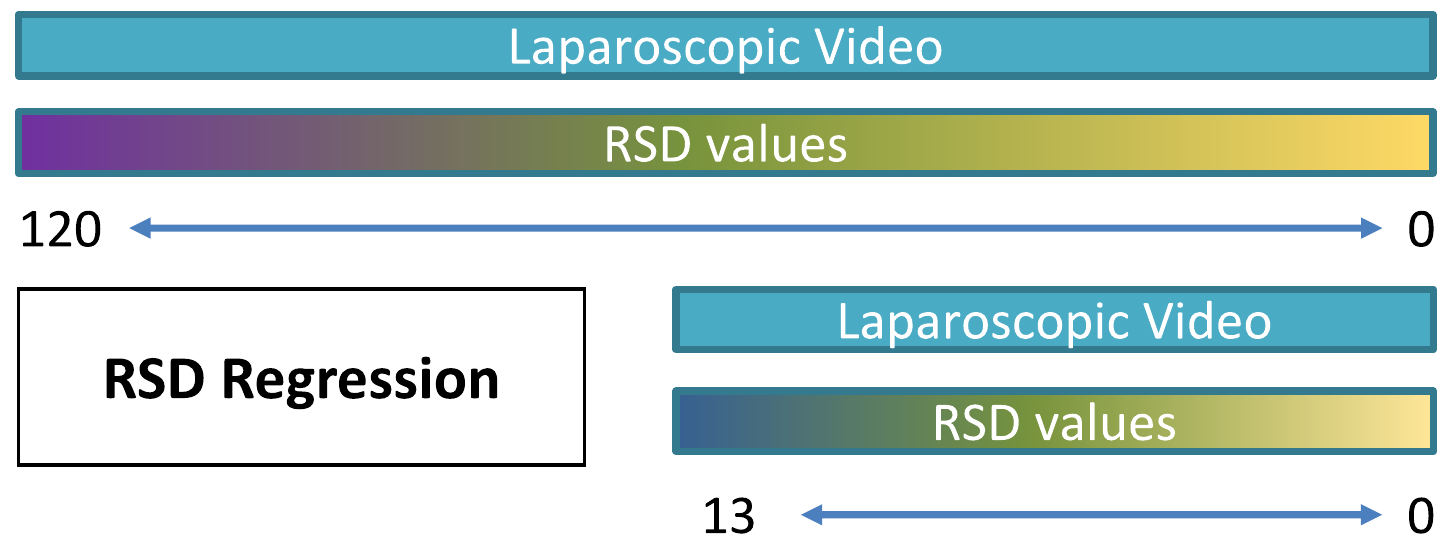}\caption{}
\end{subfigure}
\begin{subfigure}[t]{0.49\textwidth}
\includegraphics[width = 8cm]
{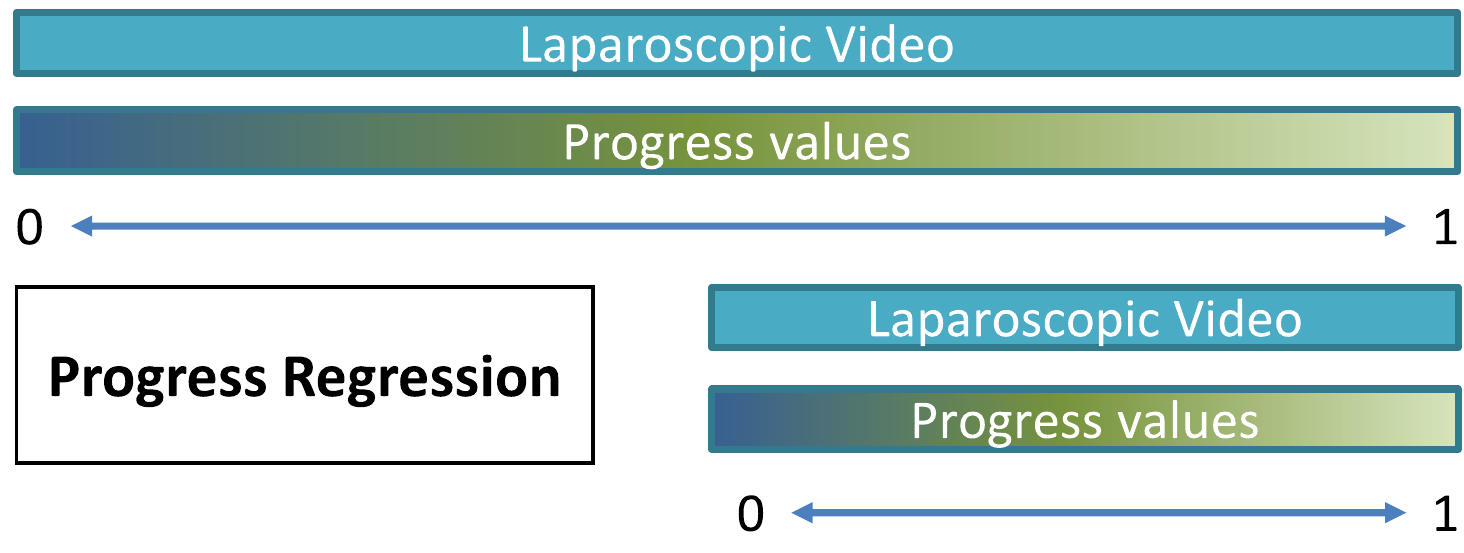}\caption{}
\end{subfigure}
\caption{Illustration of four duration-related task formulations for CNN training. (a) and (b) illustrate the formulation of RSD and progress prediction, respectively, as classification tasks. (c) and (d) illustrate their formulation as regression tasks.} \label{fig:illutration-task-formulation}
\end{figure*}

One of the main objectives of this paper is to remove the need for any labels that are acquired from manual annotation. Specifically, we would like to remove the need for phase labels from the CNN training of the pipeline proposed in our previous work \cite{aksamentov2017miccai}. Since duration-related labels are the only labels that can be automatically generated, it is then required to design a duration-related task to be performed by the CNN. This is however not an easy task since CNN works in a frame-wise manner and processes spatial features, while duration-related task highly depends on temporal information. In order to achieve this goal, we have explored various formulations of the duration-related tasks using the Cholec120 dataset. The illustration of the task formulations is shown in Fig. \ref{fig:illutration-task-formulation}.

\miniTitle{RSD classification.} This task is formulated by dividing videos into bins of 3 minutes, i.e., bin-1 contains all frames from the last 3 minutes of the video, bin-2 contains all frames from 3 minutes before bin-1, etc. We set the maximum number of bins to 20 to make the problem bounded. The CNN is then finetuned to perform RSD classification. However, the best accuracy obtained from the validation subset of a fold is around 15\%, which is very low. This could be due to the fact that there is an imbalance in the dataset, i.e., all videos contribute frames to bin-1, while only a few contribute to bin-20 (as depicted in Fig. \ref{fig:illutration-task-formulation}).

\miniTitle{Progress classification.} To mitigate the imbalance of the dataset, we decided to normalize the duration labels. We do so by consistently splitting the videos into 10 classes, as shown in Fig. \ref{fig:illutration-task-formulation}. By doing so, the task essentially becomes \textit{progress classification}. However, despite the balance in the dataset, we still observe a low accuracy of 30\% from the validation subset. This is probably due to the fact that we are addressing regression problem as classification, causing improper penalization for misclassification, i.e., an equal loss is applied if bin-1 is misclassified either as bin-2 or as bin-10, while in this particular case, misclassification as bin-10 should be penalized more.

\miniTitle{RSD regression.} Here, we finetune the CNN to directly regress the RSD values (in minutes). However, the training does not reach convergence and this could be due to a couple of shortcomings of this formulation: (1) the CNN completely neglects temporal information and (2) frames with similar regression target values are unlikely to possess similar visual or spatial characteristics. For instance, frames with $t_{rsd}=30$ mins from different videos might contain significantly different scenes due to various factors, such as surgeons' skills and patients' conditions. Hence, it is difficult for only a CNN to perform this regression task.

\miniTitle{Progress regression.} Ultimately, designing progress regression as the task that is carried out by the CNN yields the best results. We believe this could be due to two reasons. First, it removes the improper penalization inherent in the classification formulations. Second, progress formulation promotes similarities between frames with similar regression target values unlike the RSD formulation. For example, frames with $prog=0.5$ from different videos are likely to contain similar scenes since this is the middle point of the surgery and with a consistent workflow, it should fall on more or less the same task for every surgery (of the same type). \\
In all folds, the CNN training converges, yielding low MAE of progress estimation on the validation subset: $\sim$15\%. One might suggest that considering the low MAE of progress estimation, the CNN should be sufficient to perform RSD estimation. However, as seen earlier, indirectly deriving RSD estimates from progress estimates using Eq. \ref{eq:derived-rsd} does not provide optimal performance and results in a high MAE of $\sim$11 min on the validation subset.

\subsection{Importance of CNN Finetuning}

For all deep methods, the first optimization process consists of finetuning ResNet, either on phase recognition or progress regression. To evaluate the importance of finetuning, we remove the CNN training process from the two-step optimization of {\ours}, i.e., we use a ResNet model which has been pre-trained on ImageNet to extract the visual features from the video frames.

On Cholec120 this pipeline yields an MAE of 12.3$\pm$6.4 mins, which is significantly worse than {\ours}. This indicates the importance of CNN finetuning to obtain desirable results. This is however expected, since here the ResNet is only trained on ImageNet dataset which contains images that are substantially different to the images in our dataset. It is also interesting to observe that despite the complex deep model behind this pipeline, it still performs generally worse than the Na\"ive approach. This might also be correlated to the fact that we pass \textit{sub-optimal} visual features extracted by the non-finetuned ResNet to the LSTM, making the LSTM training difficult to converge.

\subsection{LSTM Cell Visualization}

\begin{figure}[t]
\begin{centering}
\includegraphics[width=8cm]{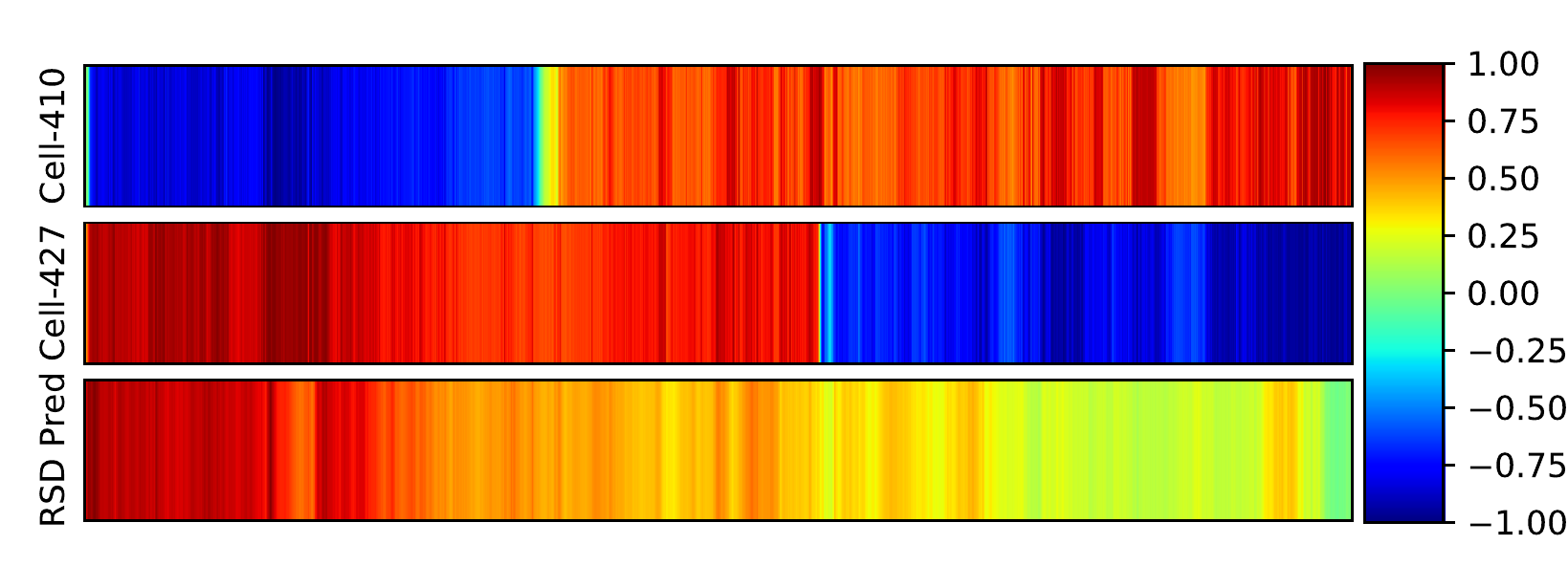}
\par\end{centering}
\caption{Activations of cell-410 and cell-427, and the normalized RSD prediction from the network. The x-axis represents the temporal axis, where the left-most is the beginning of video and the right-most is the end of the video. (Best seen in color) \label{fig:increase-decrease}}
\end{figure}
We have shown that the deep learning pipeline with the LSTM network works well in performing RSD estimation. To further understand what types of features are learnt by the LSTM, we here perform an analysis similar to the one in \cite{karpathy2015arxiv}. Specifically, we go through the LSTM cells, analyze the activation pattern and correlate it to the visual evidence from the video. The analysis is performed on videos from the subset E of Cholec120 using \ours. 

The first interesting pattern that we found in the cells is that there are several cells containing decreasing/increasing values. Examples of such cells are shown in Fig. \ref{fig:increase-decrease}, where we can observe that the activation values of cell-410 of the LSTM increases as the surgery goes on and the activation values of cell-427 decreases. These cells essentially mimic the evolution of surgery progress and remaining surgery duration, respectively. Intuitively, the presence of such cells is important to accurately predict the RSD.

\begin{figure}[t]
\begin{centering}
\includegraphics[width=7.5cm]{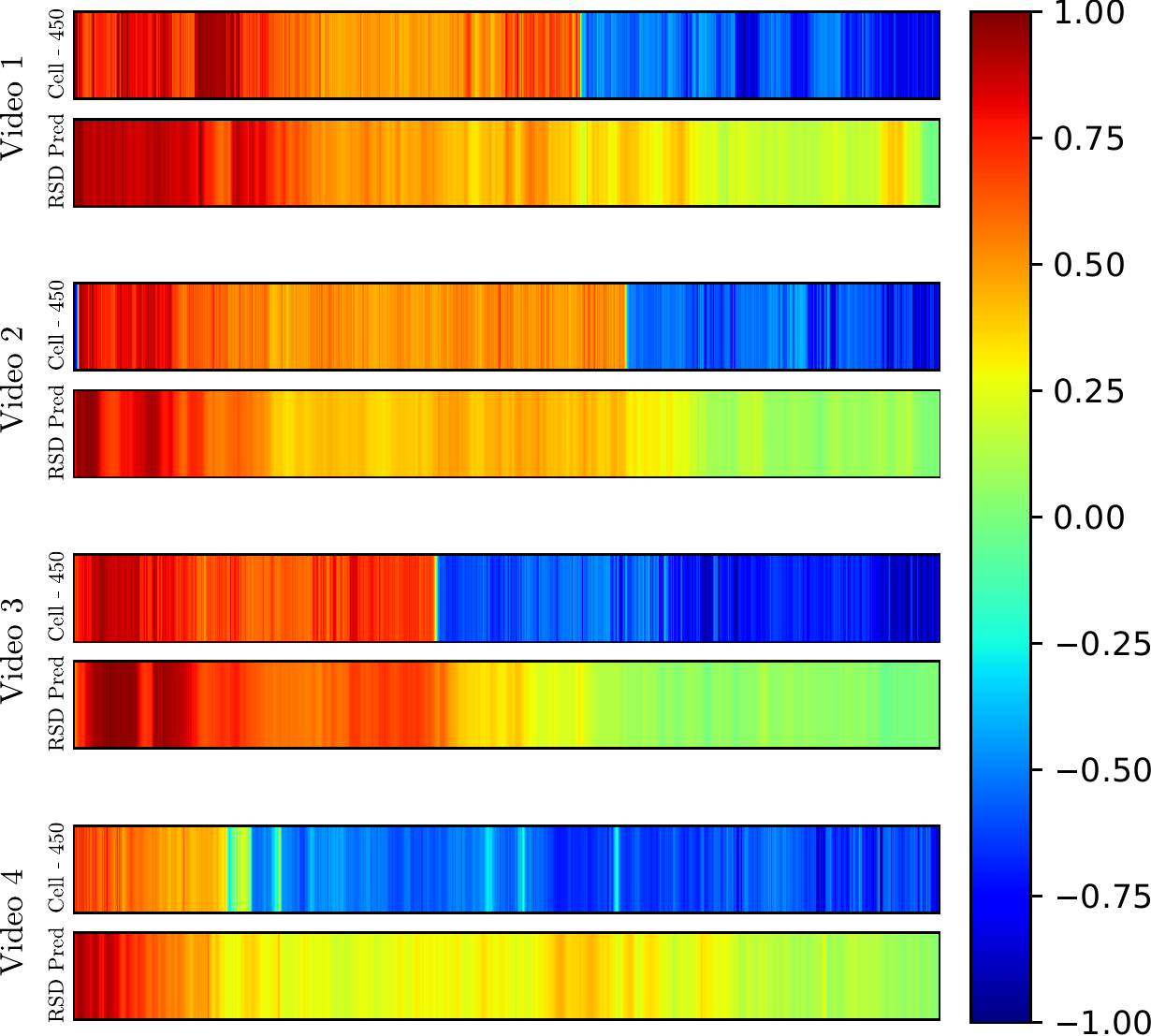}
\par\end{centering}
\caption{Cell-450 activation (top) and normalized RSD prediction (bottom) from four different videos. The x-axis represents the temporal axis, where the left-most is the beginning of video and the right-most is the end of the video. (Best seen in color)\label{fig:halves-activation}}
\end{figure}

We have also found that there are certain cells in which the activation patterns split the videos into two parts. For example, we show the activations of cell-450 from four different videos in Fig. \ref{fig:halves-activation}. After comparing the cell activation with the RSD estimation, this cell plays an important role in deciding at which point 50\% progress has been achieved. This is particularly interesting, because, due to variation in surgical performance, there is no specific visual cue why a certain point in time during a surgery is predicted by the cell as the half point: in the four videos, the half points occur in different surgical phases and in the presence of different surgical tools.

\begin{figure}[t]
\begin{centering}
\begin{subfigure}[t]{0.45\textwidth}
\includegraphics[width = 8cm]{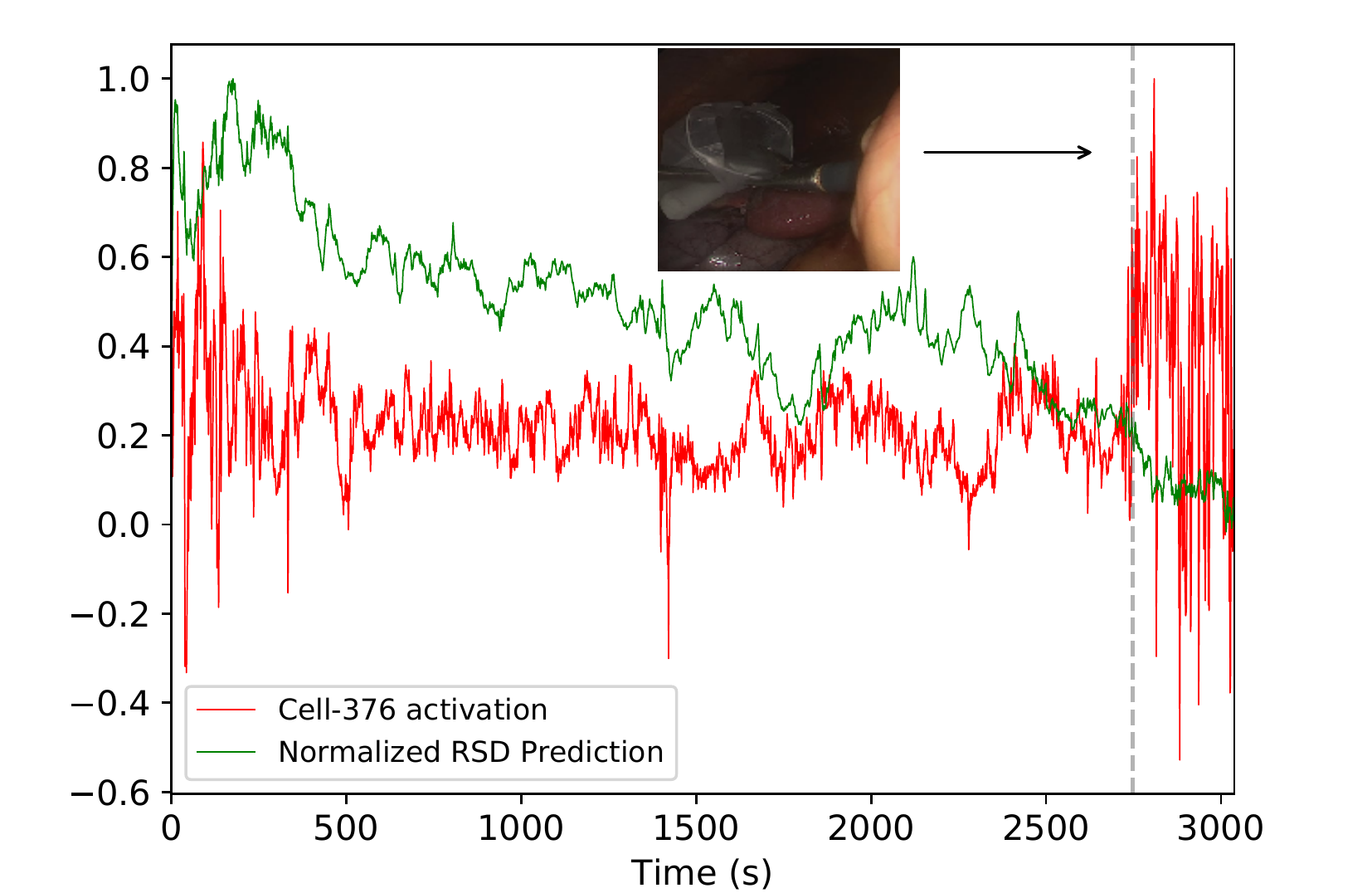}
\end{subfigure}
\begin{subfigure}[t]{0.45\textwidth}
\includegraphics[width = 8.1cm]{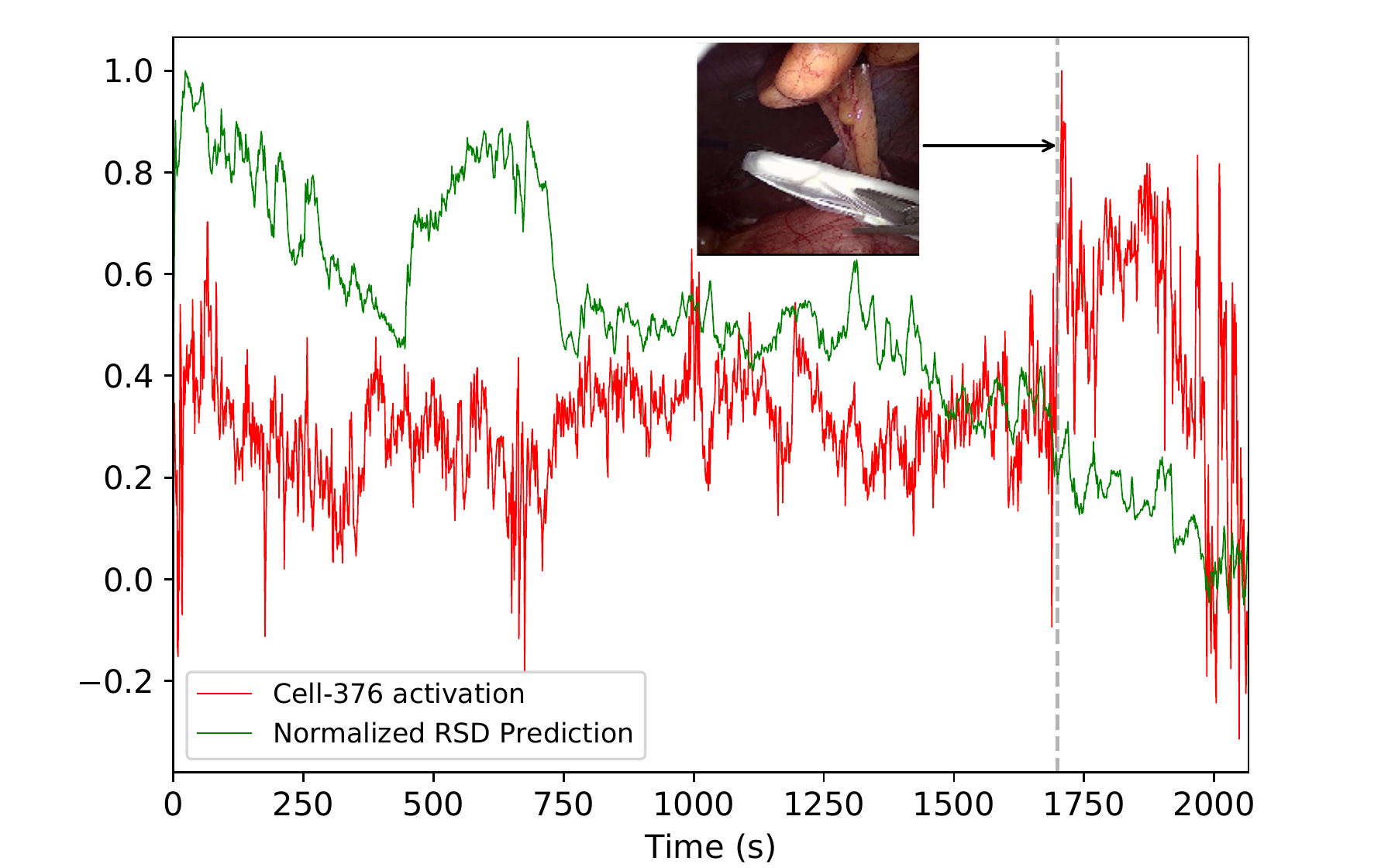}
\end{subfigure}
\par\end{centering}
\caption{Graphs of cell-376 activation and normalized RSD estimation for two different videos. \label{fig:cell6activation}}
\end{figure}

Another interesting behavior that we observed is that several cells (e.g., cell-376) respond strongly to the presence of the specimen bag. We show the cell activation and the normalized RSD graphs for two different cholecystectomy videos in Fig. \ref{fig:cell6activation}. It can clearly be seen that when a specimen bag is in the frame, the cell activation varies significantly, which also drops the RSD estimation. This shows that the LSTM network learns that the surgery is nearing to the end when a specimen bag appears.

\section{Conclusions}

In this paper, we have proposed a deep learning pipeline, referred to as {\ours}, to estimate remaining surgery duration solely relying on the visual information from laparoscopic videos. The main advantage of {\ours} is that its training process does not require any labels obtained from manual annotation. Instead, the labels are automatically generated from the dataset. On a large dataset containing 120 cholecystectomy videos, we have shown that {\ours} yields a performance comparable to an approach which requires phase labels from an expert at training time. The fact that {\ours} does not require any manual annotations makes it easily adaptable to perform RSD estimation on other surgeries. In this paper, we have shown that {\ours} is generalizable to bypass surgeries (tested on a dataset of 170 videos) and thus has the potential to be used in other laparoscopic surgeries. With such an approach, a better OR management system, which can intraoperatively predict all the remaining surgery durations inside a surgical facility, could be built. Subsequently, the throughput of the OR could be significantly optimized. In addition, an optimal management system based on {\ours} would promote patients' comfort and safety by reducing waiting time and duration of anesthesia.

One drawback of the proposed approach is that the complete pipeline is trained in a two-step manner. As a result, the loss from the LSTM network is not backpropagated to the CNN, resulting in the lack of temporal context in the visual features extracted by the CNN. In the future work, it would be interesting to implement a method which enables the complete pipeline training in an end-to-end manner. The pipeline also relies solely on visual information during the procedure. Available preoperative information, such as patient condition, and surgeon's skills, could also be incorporated explicitly into the pipeline as additional features to make a better prediction model. We also plan to deploy RSDNet in the OR and evaluate the potential benefits that it provides.

\bibliographystyle{IEEEtran}

\end{document}